\documentclass[conference]{IEEEtran}

\usepackage{booktabs}
\usepackage{graphics} %
\usepackage{epsfig} %
\usepackage{mathptmx} %
\usepackage{times} %
\usepackage{amsmath} %
\usepackage{amssymb}  %
\usepackage{arydshln} %
\usepackage{capt-of}
\usepackage{textcomp}  %
\usepackage[dvipsnames]{xcolor}
\usepackage{xcolor,colortbl}
\usepackage[accsupp]{axessibility}  %
\usepackage{multirow}
\usepackage{array}
\let\labelindent\relax
\usepackage[shortlabels]{enumitem} %
\makeatletter
\let\NAT@parse\undefined
\makeatother
\definecolor{citepurple}{rgb}{0.288,0.1196,0.7}
\usepackage[backref=page,breaklinks,colorlinks,bookmarks,citecolor=citepurple]{hyperref}
\usepackage{algorithm}
\usepackage{duckuments}
\usepackage{algpseudocode}

\definecolor{Gray}{gray}{0.90}
\newcolumntype{g}{>{\columncolor{Gray}}c}
\definecolor{ffe1da}{RGB}{255,225,218}
\definecolor{F7E0D5}{RGB}{247,224,213}
\definecolor{darkF7E0D5}{RGB}{209,154,128}
\colorlet{Light}{White!0!F7E0D5}
\definecolor{bleudefrance}{rgb}{0.19, 0.55, 0.91}

\newcommand{\coolname}{\textit{ViSafe}}
\newcommand{\coolnamenoit}{ViSafe}

\definecolor{darkpurple}{rgb}{0.288,0.1196,0.7}

\definecolor{amber}{rgb}{1.0, 0.75, 0.0}

\newcommand{\etal}{\emph{et al.}}

\definecolor{darkgray}{rgb}{0.2, 0.2, 0.2}
\newcommand{\highlight}[1]{\textcolor{darkgray}{\textbf{#1}}}

\usepackage[capitalize]{cleveref}
\crefname{section}{Section}{Sections}
\crefname{table}{Table}{Tables}

\makeatletter
\newcommand{\cdashmidrule}[1]{%
  \noalign{\vskip\aboverulesep}
  \cdashline{#1}
  \noalign{\vskip\belowrulesep}}
\makeatother

\newcommand*{\subfigref}[2][]{%
  Fig. \hyperref[{fig:#2}]{%
    \ref*{fig:#2}%
    \ifx\\#1\\%
    \else
      \,#1%
    \fi
  }%
}

\usepackage[numbers]{natbib}
\usepackage{multicol}

\newcommand{\webpage}{https://theairlab.org/visafe/}
\newcommand{\webtext}{\color{orange}{\textbf{theairlab.org/visafe/}}}

\newcommand{\authorhref}[3][citepurple]{\href{#2}{\color{#1}{#3}}}

\begin{document}

\title{
Demonstrating \coolnamenoit: Vision-enabled Safety for \\
High-speed Detect and Avoid
\\[6pt]
\Large{\href{\webpage}{\webtext}}
\vspace{-1em}
}

\author{
\authorhref{https://parvkpr.github.io/}{Parv Kapoor}$^{*}$,
\authorhref{https://www.linkedin.com/in/ian-higgins-53957718a}{Ian Higgins}$^{*}$,
\authorhref{https://nik-v9.github.io/}{Nikhil Keetha}$^{*}$,
\authorhref{https://www.jaypatrikar.me/}{Jay Patrikar}$^{*}$,
\authorhref{https://www.bradymoon.com/}{Brady Moon},
\authorhref{https://www.linkedin.com/in/zelinye}{Zelin Ye},
\authorhref{https://shockwaveHe.github.io/}{Yao He},
\\
\authorhref{https://www.ivancisneros.com/}{Ivan Cisneros},
\authorhref{http://www.huyaoyu.com/}{Yaoyu Hu},
\authorhref{https://www.cs.cmu.edu/~cliu6/}{Changliu Liu},
\authorhref{https://eskang.github.io/}{Eunsuk Kang},
\authorhref{https://theairlab.org/team/sebastian/}{Sebastian Scherer}
\\[5 pt]
\href{https://www.ri.cmu.edu/}{Carnegie Mellon University}
\\[5 pt]
*Equal Contribution 
}

\makeatletter
\let\@oldmaketitle\@maketitle
\renewcommand{\@maketitle}{\@oldmaketitle
\centering
\begin{tabular}{cccc}
\includegraphics[width=.99\textwidth]{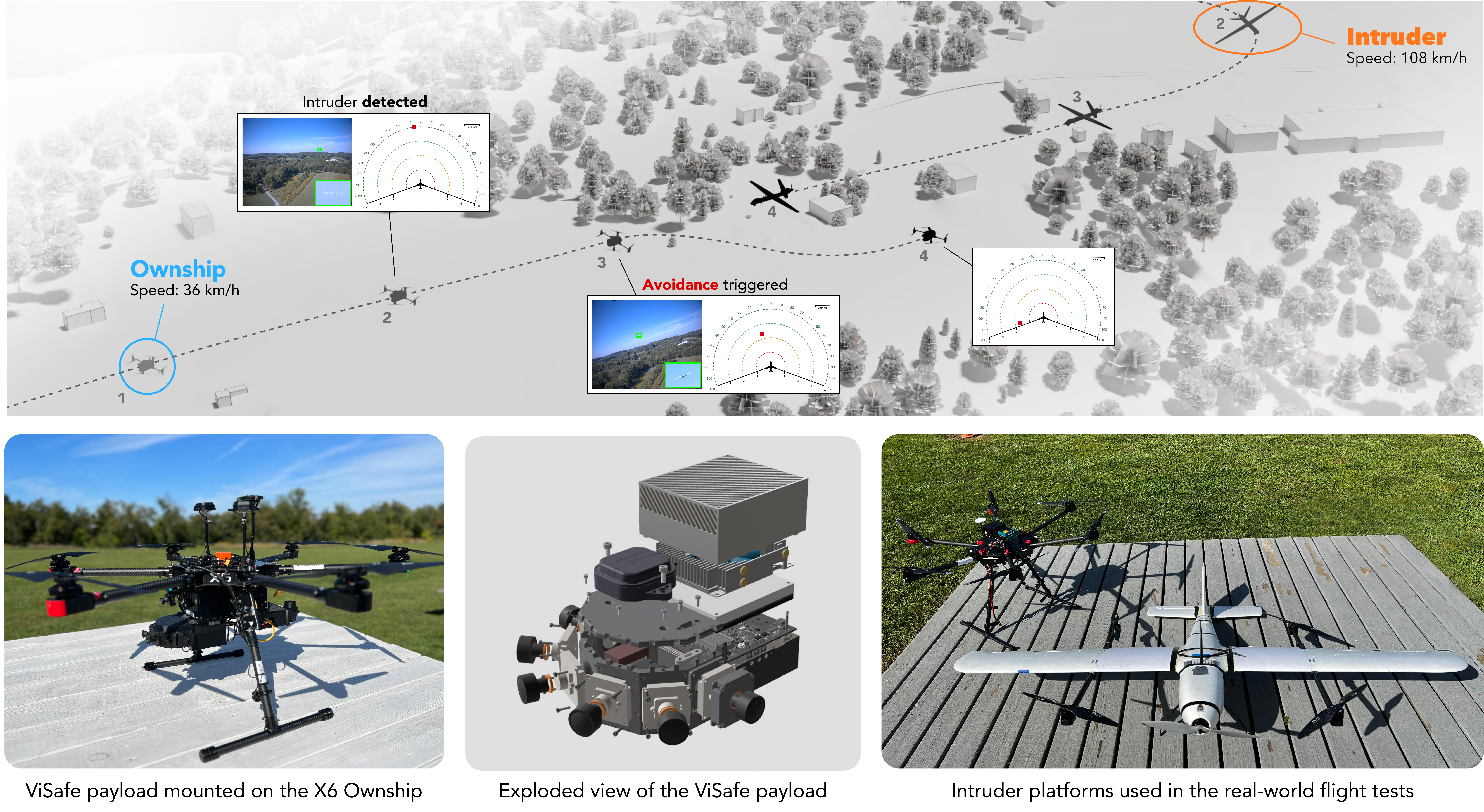}
\end{tabular}
\captionof{figure}{\textbf{We demonstrate \coolname{}, a high-speed vision-only airborne collision avoidance system.} \textit{Top:} Rendering of a real-world flight test log where the \coolname{} system detects an incoming intruder with a \highlight{144 km/h closure rate} and performs an avoidance maneuver to ensure safe separation. The annotations showcase the detections and multi-camera tracks from the vision-based aircraft detection and tracking system, while the trajectory shows the log of the performed real-world avoidance maneuver. The numbered annotations showcase the different stages of the flight test, from intruder detection to avoidance completion. \textit{Bottom:} ViSafe payload, ownship, and intruder platforms used in the real-world flight tests.
}
\label{fig:splash}
}
\makeatother

\maketitle

\begin{abstract}
Assured safe-separation is essential for achieving seamless high-density operation of airborne vehicles in a shared airspace.
To equip resource-constrained aerial systems with this safety-critical capability, we present \coolname{}, a high-speed vision-only airborne collision avoidance system. 
\coolname{} offers a full-stack solution to the Detect and Avoid (DAA) problem by tightly integrating a learning-based edge-AI framework with a custom multi-camera hardware prototype designed under SWaP-C constraints.
By leveraging perceptual input-focused control barrier functions (CBF) to design, encode, and enforce safety thresholds, \coolname{} can provide provably safe runtime guarantees for self-separation in high-speed aerial operations.
We evaluate \coolname{}’s performance through an extensive test campaign involving both simulated digital twins and real-world flight scenarios. 
By independently varying agent types, closure rates, interaction geometries, and environmental conditions (e.g., weather and lighting), we demonstrate that \coolname{} consistently ensures self-separation across diverse scenarios. 
In first-of-its-kind real-world high-speed collision avoidance tests with closure rates reaching \textit{144 km/h}, \coolname{} sets a new benchmark for vision-only autonomous collision avoidance, establishing a new standard for safety in high-speed aerial navigation.

\end{abstract}

\IEEEpeerreviewmaketitle

\setcounter{figure}{1} %

\section{Introduction}
\label{sec:intro}

Collision avoidance systems are critical to enabling safe operations in shared airspace. 
The integration of Uncrewed Aerial Systems (UASs) into an already congested National Airspace System raises pressing concerns about ensuring the safe separation of airborne vehicles.
Existing solutions, such as Autonomous Collision Avoidance Systems (ACAS)~\cite{7778055} and Unmanned Traffic Management (UTM) \cite{hamissi2024comprehensive} frameworks, have demonstrated effectiveness. 
However, these systems often depend on multiple active sensor modalities—such as transponders, radars, and ADS-B—which are unsuitable for small UASs due to stringent size, weight, power, and cost (SWaP-C) constraints. 
Addressing threats posed by cooperative as well as non-cooperative aerial entities like balloons and rogue drones while respecting SWaP-C resource constraints remains an ongoing challenge. 
As a result, current regulations impose strict line-of-sight requirements on human operators, significantly restricting the utility and scalability of UASs.

Cameras offer a lightweight, cost-effective alternative for enabling safety-critical Detect and Avoid (DAA) capabilities in sUAS. With the growth of data-driven methods, vision-based object detection offers a promising direction for tracking small aircraft in images with low signal-to-noise ratios \cite{ghosh2023airtrack}. However, real-world deployment and integration of these detection systems with downstream provably safe collision avoidance methods remain challenging.

We present \coolname{}, a vision-only airborne collision avoidance system to impart see-and-avoid capabilities to sUAS. \coolname{} builds on our prior work AirTrack \cite{ghosh2023airtrack}, which uses high-resolution detection and tracking networks to detect aerial objects. We extend AirTrack to the multi-camera setting by using multi-view fusion to track detected intruder positions across multiple cameras. Additionally, we formulate the downstream vision-based collision avoidance problem from a control theoretic perspective using Control Barrier Functions (CBFs) \cite{8796030}, which provide provable guarantees of safety and are suited for runtime monitoring and response.

CBFs have been successfully applied in various domains, including safe navigation \cite{harms2024neuralcontrolbarrierfunctions, Agrawal2017DiscreteCB}, robotic manipulation \cite{yu2024efficientmotionplanningmanipulators, 9636794}, and industrial automation systems \cite{8796030}. Their deployment for safe collision avoidance in airspaces has also been investigated \cite{10327509, molnar2024collisionavoidancegeofencingfixedwing, patrikar2022challenges}. 
While existing work assumes global availability of information and uses proprioceptive information, \coolname{} removes this assumption in the formulation. 
We identify key challenges in deploying CBFs in the wild and provide insights for deploying these techniques for high-speed collision avoidance. 
Moreover, we offer an edge compute-focused solution for real-time deployment on resource-constrained platforms.

Overall, we adopt a ``requirement refinement" approach for designing our formulations, where we leverage the empirical performance statistics of our vision inference system (AirTrack~\cite{ghosh2023airtrack}) to account for uncertainty in state estimation. This is achieved using a multi-view Kalman Filter and the ASTM F3442/F3442M standard (satisfied by AirTrack) to determine the range profile in which intruder detections are reliable.

Bridging the gap between theoretical formulations and real-world safe behavior requires extensive testing. To address this, we propose a ``digital twin" of the system and the operating environment in simulation. Using Nvidia Isaac Sim \cite{isaac_sim_ref}, we render realistic camera feeds for use in our visual detection algorithm, providing large-scale and diverse scenario benchmarking abilities. These tests were run using the same hardware as the real-world payload, thereby minimizing our sim-to-real gap for testing.

Finally, after extensive experimentation in simulation, we test our avoidance algorithms at multiple real-world testing facilities. We created representative configurations in the field to analyze our system's collision avoidance capabilities. We conducted approximately 80 hours of flight testing across two outdoor testing locations to validate our hypotheses. We observe similar collision avoidance performance in both simulation and field testing, which further lends credibility to our hardware-in-the-loop simulation fidelity.

The main contributions of this work are as follows:
\begin{enumerate}
    \item Multi-view vision-only aircraft detection \& tracking and CBF-based collision avoidance system that assumes no global availability of information or communication.
    \item Custom-built SWaP-C hardware that simultaneously streams multiple camera inputs, provides state estimation, performs deep learning model edge inference, and computes avoidance maneuvers on board in real time.
    \item Digital twin and hardware-in-the-loop simulation to perform DAA benchmarking and performance analysis under different agent types, closure rates, interaction geometries, and environmental conditions.
    \item First-of-its-kind real-world flight tests demonstrating that \coolname{} ensures safe aerial separation in encounter scenarios with a closure rate of up to 144 km/h.
\end{enumerate}

\section{Related Work}
\label{sec:related}
\subsection{Collision Avoidance logics}
One of the seminal works in airborne collision avoidance was the Traffic Collision and Avoidance System (TCAS) \cite{kuchar2007traffic}, which functions on a cooperative surveillance mechanism. The agents communicate their position and intent using Mode S transponders, which must be installed on every aircraft. More recently, the ACAS family of avoidance algorithms has been developed. There are variants of this algorithm for different agent types in different airspaces (ACAS Xa, Xu), etc. The key factor driving the development of ACAS algorithms is the availability of extended surveillance data using ADS-B, which enables aircraft to assess potential collision risks and coordinate maneuvers collaboratively. These logics involve generating cost tables for agent states and possible actions through simulation and optimization \cite{asmar2013optimized}. These tables are then used for Resolution Advisories (RAs) during deployment.

However, most existing avoidance logics require special sensors and information to provide RAs. Deploying these sensors on small UAS is challenging, and there is a large push toward sensor miniaturization for this purpose. The investigation of ACASXu with visual inference information is a relatively unexplored area \cite{6875652}. Additionally, most avoidance logics are designed for large UAS, and hence, resolution advisories tend to be relatively simple. However, small UAS can be in dense airspaces with other UAS where a finer discretization of action space is needed. This can lead to state space explosion, where cost tables can be prohibitively large. 

Our proposed technique works with vision inference data and does not depend on radar and ADS-B data. Additionally, our RAs are generated at runtime using efficient mixed integer linear programming (MILP) solvers, which are suited for more fine-grained control of UAS. Moreover, our technique can factor in nominal control inputs, which incorporate the liveness requirement of reaching a goal. It helps our agent stray not too far from the originally planned trajectory.

\subsection{Control Barrier Functions for Aerial Collision Avoidance}

There has been a recent surge in using CBFs to ensure the safe separation of UAS. Squires \etal \cite{8511342} identify key challenges with designing CBFs for collision avoidance and propose a construction technique. However, they assume information availability at all times during an episode and test their formulations in simulation without sensor noise. Likewise, follow-up work \cite{DBLP:journals/corr/abs-1906-03771} investigates the deployment of CBFs in a decentralized setup with message passing to communicate control outputs. However, they do not consider non-cooperative setups in which all agents do not provide information. Our proposed technique assumes a non-cooperative setup in which no information about other agents is explicitly communicated to the controlled agent. Moreover, we deal with information unavailability, where the CBF is only active when intruder information from our vision-based detection and tracking module is available.

\subsection{Aircraft Detection and Tracking}

Historically, traditional vision-based aircraft detection and tracking systems have employed modular approaches incorporating established computer vision techniques such as frame stabilization, background-foreground separation, and flow-based object tracking~\cite{lai2013characterization, dey2010passive, fasano2014morphological, carnie2006image, lai2011airborne, mejias2010vision}. 
In these modular approaches, frame stabilization was achieved using optical flow or image registration methods~\cite{mccandless1999detection, reilly2010detection, schubert2014robust}, while regression-based motion compensation and morphological operations were utilized to highlight potential intruders~\cite{rozantsev2015flying}. 
Background subtraction and traditional machine learning algorithms, including SVMs, were also applied to learn descriptors to identify aircraft~\cite{rozantsev2015flying, dey2010passive, petridis2008learning}. 
Furthermore, temporal filtering was used to reduce the false positives from these morphological operations.
Lastly, track-before-detect methods have also been explored, where the tracking is usually managed using methods like Hidden Markov Models, Kalman filters, and Viterbi-based filtering~\cite{nussberger2014aerial, lai2013characterization, lai2008hidden, molloy2017detection}.
While our visual detection and tracking system, AirTrack~\cite{ghosh2023airtrack}, uses deep learning-based modules, the overall sequential pipeline of performing frame alignment, detection, tracking, and false positive filtering is strongly motivated by the success of past modular \& traditional approaches.

Recent advancements have leveraged deep learning, particularly convolutional neural networks (CNNs), for object detection~\cite{chen2014aircraft, hwang2018aircraft, stojnic2021method, james2018learning}. 
In particular, for aircraft detection, the challenge is detecting small objects within high-resolution images, where keypoint-based architectures prove more effective than traditional anchor-based methods like YOLO and R-CNN~\cite{duan2019centernet}. Given the data-centric nature of these methods, this has led to a proliferation of aircraft detection datasets \cite{aircrowd,smyers2023avoidds,shi2023complex,yang2024robust,patrikar2025image}.
This insight aligns with findings in related fields such as face detection and object detection in aerial imagery~\cite{hu2017finding, keetha2022airobject, liang2021learning}. 
For vision-based aircraft detection systems where computational efficiency is crucial, fully convolutional networks that predict heatmaps are the most effective~\cite{james2018learning, james2019below}.
Furthermore, the tracking-by-detection framework is standard for aircraft tracking, using a management system to handle track initiation and termination and associating new detections with existing tracks using algorithms like the Hungarian method~\cite{bewley2016simple, lai2013characterization, dey2010passive}. 
For small objects, where bounding box IoU is unreliable due to sensitivity to positional deviations, integrated detection and tracking approaches offer a superior alternative~\cite{zhou2020tracking}.
Motivated by this trend, our AirTrack~\cite{ghosh2023airtrack} employs an anchor-free detection and tracking system.

\begin{figure*}
     \centering
      \includegraphics[width=.95\textwidth]{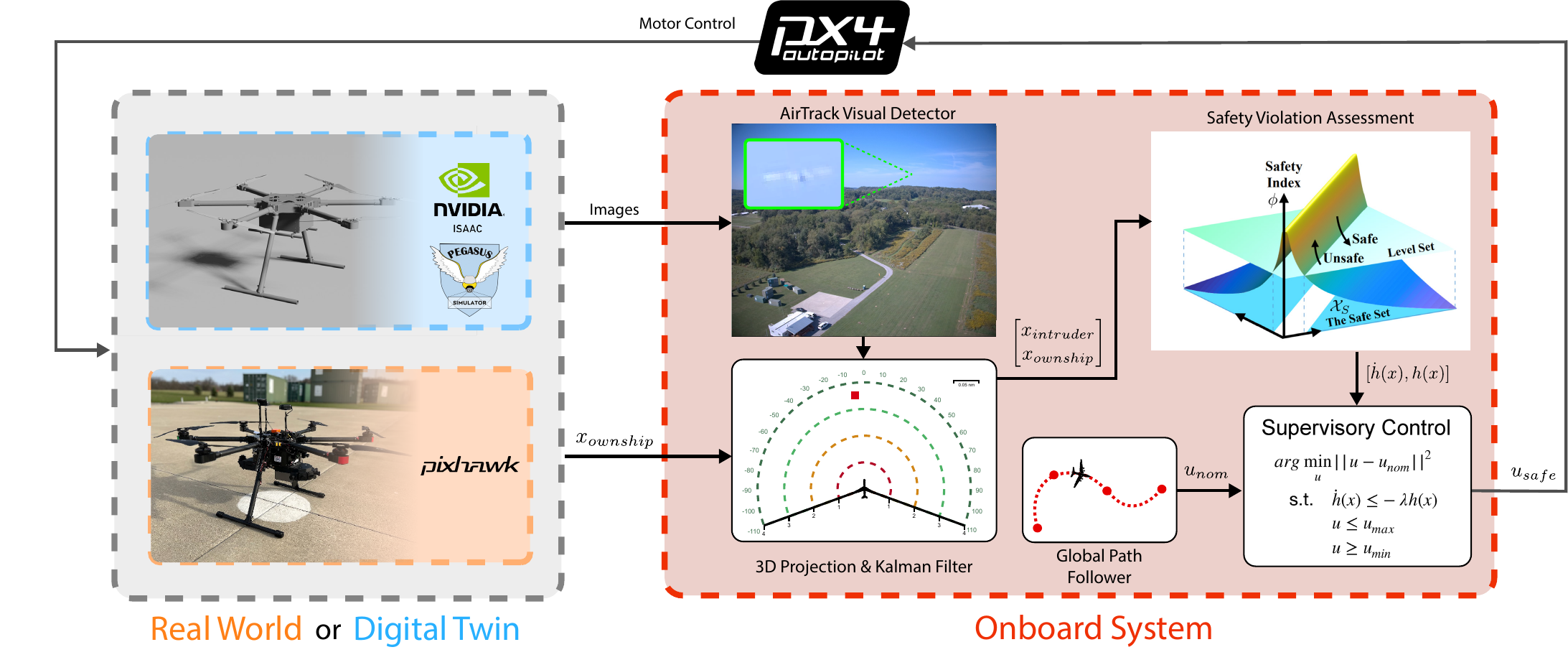}    
      \caption{
      \textbf{Overview of our \coolname{} framework for real-world testing and hardware-in-the-loop simulation:} Firstly, the onboard sensors or digital twin simulation stream the multi-cam videos to the AirTrack visual detection module, which detects the intruder across multiple views. Then, these detections, along with the ownship state information, are sent to the multi-view fusion and coordinate frame conversion module, which then tracks the intruder and sends the intruder state information along with the ownship state information to the CBF. The CBF uses the nominal global plan and the safety violation assessment to compute modifications to the nominal control input in case of violation. This safe control output is then sent to the drone autopilot system for execution. This loop continues to operate in real-time until the flight test is complete.} \label{fig:overview}
\end{figure*} 

\section{Problem Definition}
\label{sec:problem}

We model the problem of maintaining safe separation between two aerial agents: the ownship agent and the intruder agent.
The ownship refers to the primary controlled agent and the intruder refers to a potentially unknown agent in the configuration space which doesn't have any communication with the ownship agent.
Only passive sensing, specifically vision, is used to detect the intruder. 
Safe separation entails keeping the two agents outside each other's collision volume at all times. 
We model the collision avoidance problem in the horizontal plane and assume constant velocity vectors for the intruder agent inspired by \cite{7778055}. 
The separation requirement can be formally defined as:
\begin{equation} \label{eq1}
   \forall t \in T_{operation} \bullet d_{ownship-intruder} \geq d_{thresh}
\end{equation}
where $d_{thresh}$ is the minimum distance threshold and $d_{ownship-intruder}$ is the relative horizontal distance between the agents. For notational simplicity, we will refer to this distance as $d$ in our formulations. Additionally, $T_{operation}$ is the time of operation of the ownship agent, and \cref{eq1} formally states the problem of safe separation for the entire operation time.

As can be inferred from \eqref{eq1}, this is equivalent to a forward invariance property.
In order to encode and enforce this property, we make certain assumptions.
First, we assume a simple unicycle dynamics model for our ownship agent: 

\begin{equation}
\dot{x_{own}} = \begin{bmatrix} 
v_{own} \cos{(\chi_{own})} \\
v_{own} \sin{(\chi_{own})}  \\
0   \\
0 
\end{bmatrix} + 
\begin{bmatrix} 
 0 & 0  \\
0 & 0 \\
1 & 0  \\
0 & 1 
\end{bmatrix} u
\end{equation}

\begin{equation}
 u^T = [\dot{v}_{own}\ \dot{\chi}_{own}]
\end{equation}
Here, $v_{\text{own}}$ denotes the speed of the ownship, and $\chi_{\text{own}}$ is the heading (yaw) angle. The control input $u \in \mathbb{R}^2$ consists of the rate of change of speed and heading, i.e., $\dot{v}_{\text{own}}$ and $\dot{\chi}_{\text{own}}$.
Additionally, we also consider control constraints to closely mimick real world deployment scenarios.
\begin{equation*}
    u \geq u_{min}, u \leq u_{max} \ \forall t \in T_{operation} 
\end{equation*}
These control constraints are particularly important for our use case since, for some aircraft, loiter safety maneuvers are actually impossible and would lead to an aircraft crash.

Lastly, we define the dynamics of the intruder agent:

\begin{equation}
\dot{x_{int}} = \begin{bmatrix} 
v_{int} \cos{(\chi_{int})} \\
v_{int} \sin{(\chi_{int})}  \\
\end{bmatrix} 
\end{equation}
Here, $v_{\text{int}}$ denotes the speed of the intruder, and $\chi_{\text{int}}$ is the heading (yaw) angle. 
Overall, we make two assumptions: First, the intruder has constant velocity vectors as discussed earlier. Second, the intruder agent is agnostic to the ownship agent's maneuvers, i.e., there are no interaction effects. This assumption also implies that the township agent possesses no collision avoidance algorithm, since that would lead to interaction effects. These assumptions are made to keep the problem tractable but remain practical in today's airspace.

\section{\coolname{} Framework}
\label{sec:approach}

An overview of our \coolname{} framework (\cref{fig:overview}) is as follows: 
While the visual detection module (\cref{subsec:visual_detection_module}) provides an image-level intruder detection, the intruder state information is transformed to the North East Down (NED) coordinate system and broadcasted to the safety controller in its expected input format (\cref{subsec:fusion}).
The safety controller (\cref{subsec:safety_controller}) encodes the desired safe separation constraints as well as our actuation constraints and computes modifications to nominal control input in case of violation. 
These constraints are derived using our defined CBF. 
This modified control action is then converted into low-level drone commands and executed.

\subsection{Visual Detection Module}
\label{subsec:visual_detection_module}

We extend the state-of-the-art aircraft detection \& tracking model, AirTrack~\cite{ghosh2023airtrack}, to detect intruders. Specifically, we modified the AirTrack algorithm (originally focused on single-camera inference) to incorporate multi-camera inputs and enable multi-camera tracking. Furthermore, we also upgraded the SWaP-C hardware, implementing efficient multi-camera image sharing (with zero memory copies)
and further optimizing the deep learning model inference to
operate at the desired frequency required to facilitate high-speed collision avoidance on the edge. The overall design of our Detection Module comprises Frame Alignment, Detection, Secondary Classification, and Image-level Tracking. 
The model's inputs are two consecutive grayscale image frames, and it outputs a list of tracked objects with various attributes.
The model is trained on the Amazon Airborne Object Tracking (AOT) dataset~\cite{aircrowd}, and its modules are as follows:

\textit{Frame Alignment:} 
The goal of this module is to align the successive frames in a video so that the ego-camera motion can be discarded, thereby helping to distinguish foreground objects from the background. 
This is achieved by predicting the optical flow between two successive input image frames and the confidence of the predicted flow.
The input images are cropped from the center-bottom to cover high-texture details below the horizon, and a ResNet-34 backbone architecture with two prediction heads is used for alignment. 
The prediction is made at $1/32$ scale of the input, and low confidence offset predictions are rejected.

\textit{Detection:}
This module is split into two parts: a primary detection module and a secondary detection module. 
The primary module takes in two full-resolution ($1224\times1024$) grayscale image frames that have been aligned, while the secondary module receives a smaller $\approx1/5$th sized crop of the image around the top k detector outputs (based on confidence) from the primary module. 
The network is fully-convolutional (HR-Net-W32) and it produces five outputs: a center heatmap indicating the object's center, bounding box size, center offset from the grid to the object's center, track offset of the object's center from the prior frame, and object distance in log scale.

\textit{Secondary Classifier:}
A ResNet-18 module is used as a binary classifier for false-positive rejection.
The module takes cropped regions around the bounding boxes from the detection module as input and predicts whether the crop is an aircraft or not.
This improves the overall precision.

\textit{Image-level Tracking:} We use an offset tracking vector-based approach~\cite{zhou2020tracking}.
The predicted track offset vector from the detection module is subtracted from the current object center to find the location in the prior frame.
If the location matches existing tracks within a threshold, the detection is associated with an image-level track; otherwise, a new track is initialized.

Overall, the sequential modules of AirTrack work together to provide precise image-level aircraft tracks and distance from the ownship.
For \coolname{}, we run AirTrack on video streams from two cameras at 8 Hz on our Nvidia Orin AGX compute system (described in~\cref{subsec:real_exp}).
We encourage the readers to refer to the AirTrack~\cite{ghosh2023airtrack} paper for further details.

\subsection{Multi View Fusion \& Coordinate Frame Conversion}
\label{subsec:fusion}

Once AirTrack provides an image-level intruder track (from any one of the multiple cameras), we unproject the track location in the image to 3D leveraging the predicted distance of the aircraft, center of the predicted bounding box and the calibrated intrinsics of the respective camera.
Then, the 3D position of the image-level track in the camera coordinate frame is transformed to the local frame of the 3DM-GQ7 GNS/INSS module by using the multi-camera and IMU extrinsic calibration.
Since, the GNSS module provides us the heading of the ownship with respect to the North direction, we can transform the 3D image-level intruder track position to a North-East-Down (NED) coordinate frame centered about the GNSS module of the ownship.
Now given the 3D position $[x_t, y_t, z_t]$ of the image-level track, the range $d$, azimuth $\theta$ and elevation $\phi$ of the intruder in the ownship's NED coordinate frame can be calculated as:
\begin{equation}
\begin{aligned}
    & d = \sqrt{{x_t}^2 + {y_t}^2 + {z_t}^2} \\
    & \theta = \arctan\left(\frac{y_t}{x_t}\right) \\
    & \phi = -\arcsin\left(\frac{z_t}{d}\right)
\end{aligned}
\end{equation}

We leverage a multi-view fusion module to track and fuse the intruder positions across multiple cameras.
At the start, for the first image-level intruder track, we initialize two Kalman Filters to perform tracking and fusion of new observations for the same intruder.
We empirically observe that the range ($d$) is noisier than the angles ($\theta$ \& $\phi$) since the angular estimates depend solely on the 2D location
of the detection while the range is predicted by the monocular deep learning model (AirTrack). 
It is possible to use a single Kalman Filter by specifying independent process noises; however, we find that the coupling of range measurements degrades the angular estimates, requiring significant joint tuning. 
Hence, we utilize two Kalman Filters, each with a constant velocity motion model (one for angles and one for range). 
Through decoupling, the range filter accounts for high uncertainty in range while the angle filter ensures smooth tracking.
Subsequently, for each new image-level track across the different cameras, we use the angular distance to determine if the track associates with pre-existing 3D intruder tracks.
If it associates, we fuse the new observation with the two pre-existing Kalman Filters.
Otherwise, we initiate another new 3D intruder track with two Kalman Filters.
In this way, we can use Kalman Filters (unique to each intruder) to initialize, associate, and track different intruders in 3D.
Each 3D intruder track is configured to propagate the intruder position and velocity to the downstream supervisory safety controller at a frequency of 24 Hz.

\begin{figure}
     \centering
      \includegraphics[width=\columnwidth]{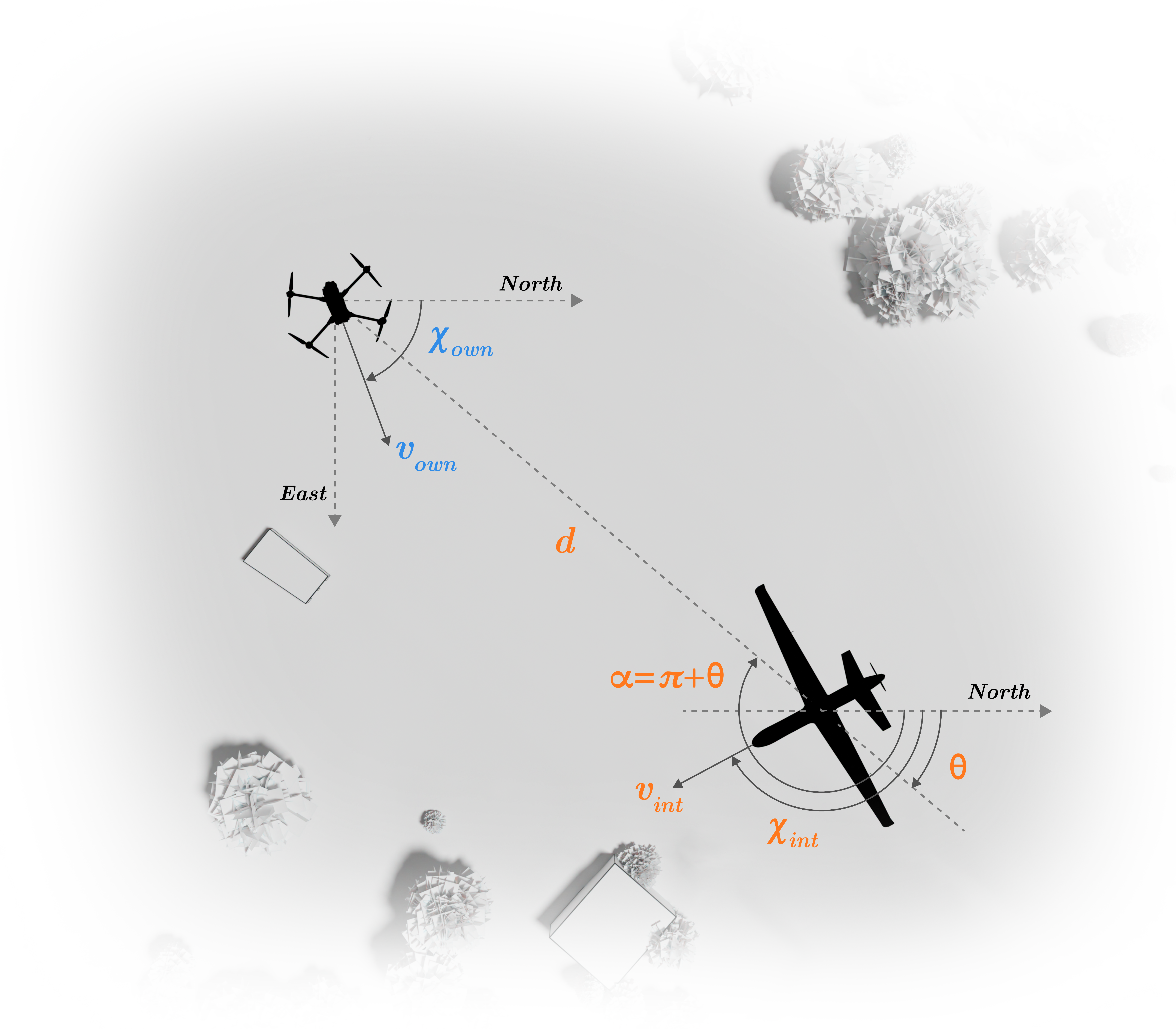}    
      \caption{\textbf{Encounter geometry and information required for vision-enabled collision avoidance.} The velocity of the ownship \textcolor{bleudefrance}{$v_{own}$} and heading with respect to North \textcolor{bleudefrance}{$\chi_{own}$} are obtained from the \textcolor{bleudefrance}{ownship odometry}. The intruder's range \textcolor{orange}{$d$}, azimuth \textcolor{orange}{$\theta$}, velocity \textcolor{orange}{$v_{int}$}, and heading with respect to North \textcolor{orange}{$\chi_{int}$} are obtained using the \textcolor{orange}{visual detection module}.}
      \label{fig:moexample}
\end{figure}

As shown in \cref{fig:moexample}, the supervisory safety controller described in the subsequent sub-section expects the multi-view fusion module to output the range $d$, the angle of the line of sight to the NED frame centered about the intruder $\alpha$, the velocity of the intruder $v_{int}$ along the North and East direction and the heading of the intruder with respect to the North direction $\chi_{int}$.
The tracks from the multi-view fusion module already provide the range $d$, and $\alpha$ can be computed as $\pi + \theta$.
Furthermore, using the ownship's odometry information ($v_{own}$ \& $\chi_{own}$) and the track information (range $d$, range rate $\dot{d}$, azimuth $\theta$ and azimuth rate $\dot{\theta}$), we can compute the two components of $v_{int}$ as follows:
\begin{equation}
\begin{aligned}
    & {v_{int}}_{North} = {v_{own}}_{North} + \dot{d}\cos{\theta} + d\dot{\theta}\sin{\theta} \\
    & {v_{int}}_{East} = {v_{own}}_{East} + \dot{d}\sin{\theta} + d\dot{\theta}\cos{\theta}
\end{aligned}
\end{equation}
Subsequently, the heading of the intruder with respect to North can be obtained as:
\begin{equation}
\begin{aligned}
    & {\chi_{int}} = \arctan\left(\frac{{v_{int}}_{East}}{{v_{int}}_{North}}\right)
\end{aligned}
\end{equation}
Hence, in this way, we transform \& convert the image-level intruder detections to the expected input format of the supervisory safety controller.

\subsection{Supervisory Safety Controller}
\label{subsec:safety_controller}

The three key requirements for this controller are:
\begin{itemize}

    \item \textbf{Certified safety}: Enforcement of a strict notion of safety while working with vision inputs 
    \item \textbf{Run time safety}: Reactive low latency solution that can make decisions in real-time 
    \item \textbf{Performant safety}: Preserving original mission requirements and introducing a minimal modification to the original control input provided by a nominal controller
\end{itemize}

Existing techniques, such as reachability analysis and regret minimization for MPC, often conflict with these three requirements. For example, reachability analysis-based techniques are certified and can generate controllers that preserve mission outputs. However, they tend to be computationally expensive, compromising the reactive requirement. Considering these three key requirements, we used a Control Barrier Function (CBF) based Quadratic Program (QP) for our supervisory safety controller.
CBFs were introduced in \cite{8796030} to formally certify the safety of robotic control systems. We assume we are given a nonlinear control affine system defined by 
\begin{equation} \label{eq0}
    \dot{x} = f(x) + g(x) u
\end{equation}
where x $\in$ $\mathcal{X}$, $f$ and $g$ are locally Lipschitz continuous functions and u $\in \mathbb{R}$  is the set of inputs to the system.

Let us say our safe set is defined by $\mathcal{X}_{s} \subset \mathcal{X}$, then the goal of safe control is to keep this set forward invariant with respect to the dynamics \cref{eq0}. Mathematically, this forward invariance of the set is described as
\begin{equation}
    \forall x(t_{0}) \in \mathcal{X}_{s} \bullet x(t) \in \mathcal{X}_{s} \quad t \geq t_0
\end{equation}

Let $\mathcal{C} := \{x \in \mathcal{X} | h(x) \leq 0\}$ be a zero sub-level set of a function $h: X \rightarrow \mathbb{R}$. Then h is a valid control barrier function for $\mathcal{C}$ given dynamics \cref{eq0} if there exists an extended class $\mathcal{K}_{\infty}$ function $\alpha : \mathbb{R} \rightarrow \mathbb{R}$ $(\alpha(0) =0) $ such that:
\begin{equation}
    L_f h(x) + L_g h(x)u \leq - \alpha(h(x) \quad \forall x \in \mathcal{C}
\end{equation}
Here, $L_f h(x)$ and $L_g h(x)$ are the Lie derivatives of $h$ along the drift $f(x)$ and control $g(x)$ dynamics, respectively. Note this formulation is slightly different than the one defined in \cite{8796030} since we define the safe set to be a zero-level subset instead of a superset. However, since we change the inequality sign, we can still ensure forward invariance using this condition. This design choice is inspired by \cite{Liu2014CONTROLIA}.

Next, we outline our CBF formulation and its associated QP-based controller. Our supervisory controller enforces our safety and actuation constraints. We devise this controller using our defined control barrier function. 
First, let our safe set be defined as $C$, then a straightforward distance maximization-based CBF would be:

\begin{equation}
h(x) = d_{thresh} - d
\label{eq:nomcbf}
\end{equation}
where 
\begin{equation*}
\begin{aligned}
x \in C : h(x) \leq 0\\
x \notin C : h(x) > 0
\end{aligned}
\end{equation*}

Here, $d$ represents the distance between the agents, and $d_{thresh}$ represents the minimum distance threshold that should be maintained between the two agents.
However, since the relative degree of pure distance maximization-based CBF with respect to $u$ is 2, there would be singularity with the CBF proposed in \cref{eq:nomcbf}. Inspired by \cite{Liu2014CONTROLIA}, we propose the following CBF:
\begin{equation}
\label{eq12}
h(x) = c + d^{n}_{thresh} - d^{n} - k\dot{d}
\end{equation}
where $c >0, n>0$ and $k>0$ are hyperparameters to be tuned.

Our defined CBF takes negative values when the agents are safely separated and positive values when the requirement is violated.  The CBF condition we enforce for forward invariance is as follows:
\begin{equation*}
   \dot{h}(x) \leq - \lambda h(x)\\
\end{equation*}

We choose the penalty term as $h(x)$ itself as is often done for CBFs. $\lambda$ is another hyperparameter that controls the rate of change of the barrier function. Note that the non-linear constraint $h(x) \leq 0$ is not necessarily a subset of $d \geq d_{thresh}$ when $\dot{d} > 0$. However, as demonstrated in \cite{Liu2014CONTROLIA}, the system remains forward invariant within the set ${h(x) \leq 0} \cap {d \geq d_{thresh}}$ under the application of the control law $\dot{h}(x) \leq - \lambda h(x)$. Our final QP formulation is defined as follows:
\begin{equation}
\begin{aligned}
arg \min_{u} \quad & ||u - u_{nom} ||^{2} \\
\textrm{s.t.} \quad & \dot{h}(x) \leq - \lambda h(x)\\
  &u \leq u_{max}    \\
    &u \geq u_{min}    \\
\end{aligned}
\label{eq:qp}
\end{equation}
where $u_{nom}$ is the control input provided by the nominal controller and $u_{max}$ and $u_{min}$ encode the actuation constraints. 

Now, to compute the constraints enforced for the QP formulation, we can concretize the various quantities in the formulation using post-processed visual detection information and ownership odometry information ($v_{own}$ \& $\chi_{own}$).
First, we derive the values of $\dot{h}$ in terms of our control input $u$ using our encounter geometry as illustrated in \cref{fig:moexample}.

\begin{equation}
    \begin{aligned}
       & \dot{h}(x) = nd^{n-1}\dot{d} -k\ddot{d} \\   
    \end{aligned}
\end{equation}
where 
\begin{equation}
    \begin{aligned}
    & \dot{d} = v_{own}\cos{(\alpha - \chi_{own})} + v_{int}\cos{(\alpha - \chi_{int})} \\  
    \end{aligned}
    \label{eq:ddot}
\end{equation}

\begin{figure*}[!t]
\scriptsize
\centering
\includegraphics[width=0.99\linewidth]{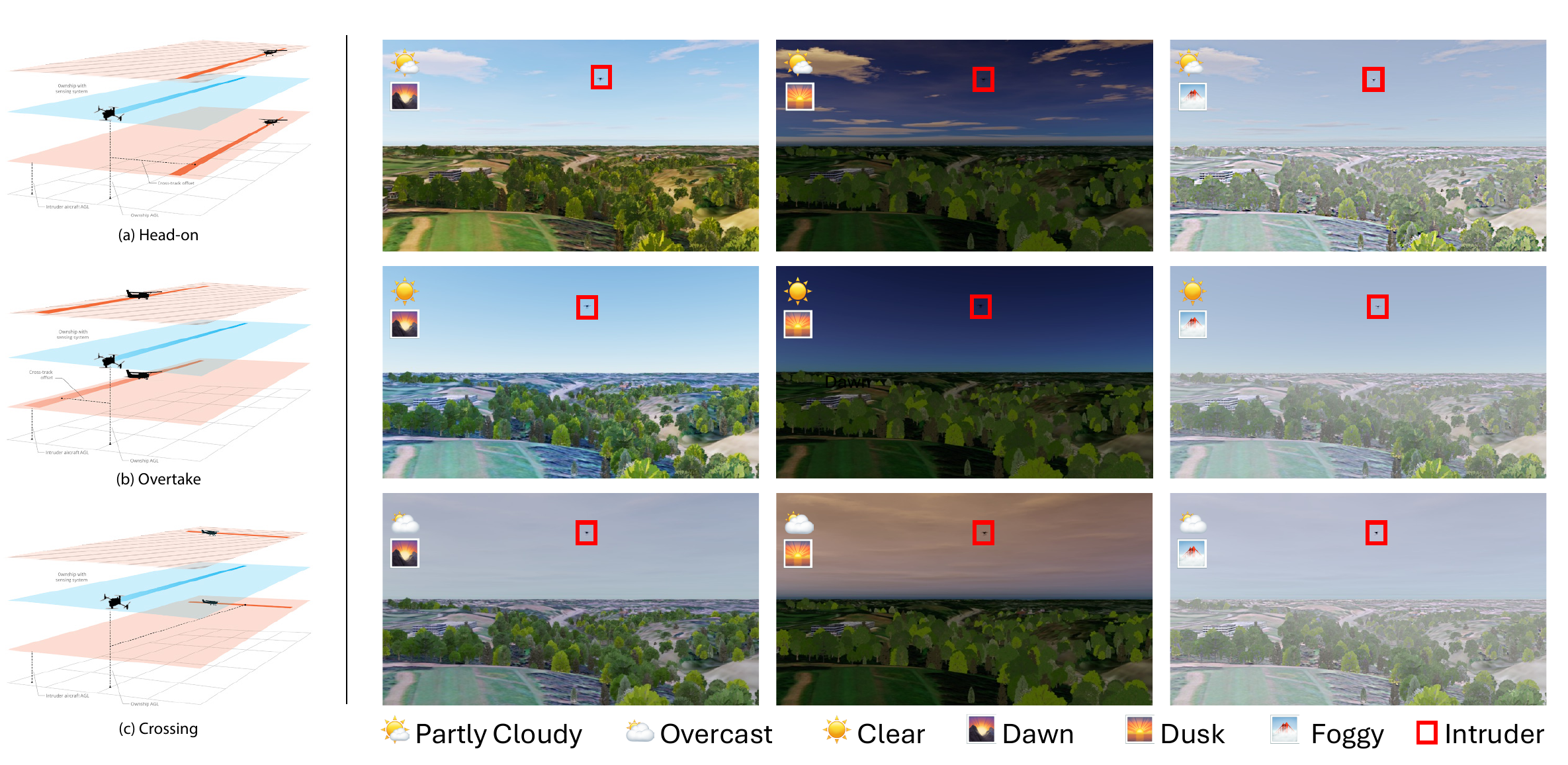}
\caption{\textbf{Diversity of the airborne collision testing scenarios:} (a) The various encounter geometries used for real-world \& simulation flight testing. (b) The diverse weather and lighting conditions that were used to evaluate \coolname{}'s robustness in simulation. In the simulation, based on the chosen encounter geometry, the intruder position is sampled randomly on the flying corridor.}
\label{fig:collisoions_and_weather}
\end{figure*}

We compute $\dot{d}$ by taking the projection of $v_{own}$ and $v_{int}$ along the line of sight (LOS). To compute $\ddot{d}$, we differentiate \cref{eq:ddot} with our original assumption of constant intruder velocity vector (i.e. $\dot{v}_{int} = 0$ and $\dot{\chi}_{int} = 0$ ) to get:
\begin{equation}
    \begin{aligned}
    & \ddot{d} = \dot{v}_{own}\cos{(\alpha - \chi_{own})} + \dot{\chi}_{own}v_{own}\sin{(\alpha - \chi_{own})} \\ 
    & + \frac{(v_{own}\sin{(\alpha - \chi_{own})}+ v_{int}\sin{(\alpha - \chi_{int})})^{2}}{d}
    \label{eq:dddot}
    \end{aligned}
\end{equation}
Now given $u =\begin{bmatrix} 
\dot{v}_{own}  \\
\dot{\chi}_{own}  \\
\end{bmatrix}$, we can rewrite \cref{eq:dddot} as
\begin{equation}
    \begin{aligned}
    & \ddot{d} = \begin{bmatrix} 
\cos{(\alpha - \chi_{own})}  \\
v_{own}\sin{(\alpha - \chi_{own})}  \\
\end{bmatrix}^\top u \\ 
    & + \frac{(v_{own}\sin{(\alpha - \chi_{own})}+ v_{int}\sin{(\alpha - \chi_{int})})^{2}}{d}
    \label{eq:dddot_rewrite}
    \end{aligned}
\end{equation}
As can be observed, the constraints on $h$ are affine in $u$ and fit the form of a QP, as defined in \cref{eq:qp}, which can be solved to compute the desired safe control $u_{safe}$.
We use the following tuned hyperparameters for the final formulation of \cref{eq12} and \ref{eq:qp}: k = 0.2, c = 0.01, n = 0.3, and $\lambda$ = 0.2.

We use a simple PD controller as our nominal controller, where the computed desired safe control $u_{safe}$ is then converted into low-level drone control actions in the form of velocity and yaw setpoints, which are executed on our ownship agent by the flight controller (PX4 in Isaac Sim and Ardupilot on the real-world drone hardware).

\begin{table*}
\centering
\caption{Experimental Configurations}
\scalebox{1}{
\begin{tabular}{ccccccc}
\toprule
\textbf{Scenario ID} & \textbf{Intruder} & \textbf{Ownship} & \textbf{Collision Geometry} & \textbf{Hor. Closure Rate} & \textbf{Location}\\

\cmidrule{1-1} \cmidrule(lr{0.75em}){2-2} \cmidrule(lr{0.75em}){3-3} \cmidrule(lr{0.75em}){4-4} \cmidrule(lr{0.75em}){5-5} \cmidrule(lr{0.75em}){6-6} 

E1& Multirotor (10 m/s) & Aurelia X6 (10 m/s) & Head-On & 20 m/s (72 km/hr) & Nardo Airfield \\
E2& Multirotor (5 m/s) & Aurelia X6 (10 m/s) & Overtake & 5 m/s (18 km/hr) & Nardo Airfield \\
E3& Multirotor (5 m/s) & Aurelia X6 (10 m/s) & Crossing & 11.18 m/s (40 km/hr) & Nardo Airfield \\
E4& VTOL (30 m/s) & Aurelia X6 (10 m/s) & Head-On & 40 m/s (144 km/hr) & Nardo Airfield \\
E5& Multirotor (10 m/s) &Aurelia X6 (10 m/s) & Head-On & 20 m/s (72 km/hr) & Leesburg, VA \\

\bottomrule
\end{tabular}
}
\label{tab:exp_setup}
\end{table*}

\begin{table*}
\centering
\caption{Digital Twin \& Hardware-in-the-Loop Benchmarking}
\scalebox{1}{
\begin{tabular}{ccccccccc}
\toprule

& \multicolumn{2}{c}{\textbf{Separation Minima (m) $\uparrow$}} & \multicolumn{2}{c}{\textbf{P(NMAC) $\downarrow$}} & \multicolumn{2}{c}{\textbf{Risk Ratio $\downarrow$}} & \multicolumn{2}{c}{\textbf{Number of violations} $\downarrow$} \\

\cmidrule{2-3} \cmidrule(lr{0.75em}){4-5} \cmidrule(lr{0.75em}){6-7} \cmidrule{8-9}

\textbf{Scenario} & Nominal & \coolname{} & Nominal & \coolname{} & Nominal & \coolname{} & Nominal & \coolname{} \\

\cmidrule{1-1} \cmidrule(lr{0.75em}){2-2} \cmidrule(lr{0.75em}){3-3} \cmidrule(lr{0.75em}){4-4} \cmidrule(lr{0.75em}){5-5} \cmidrule(lr{0.75em}){6-6} \cmidrule(lr{0.75em}){7-7} \cmidrule(lr{0.75em}){8-8} \cmidrule(lr{0.75em}){9-9}

E1 & 19.9 $\pm$ 2.15 & 35.55 $\pm$ 14.00 & 1.0 & 0.55 & 1.0 & 0.55 & 4000 & 2213  \\ 
\cdashmidrule{1-9}

E2 & 22.72 $\pm$ 2.63 & 27.635 $\pm$ 6.27 & 1.0 & 0.439 & 1.0 & 0.439 & 4000 & 1759 \\ 

\cdashmidrule{1-9}

E3 & 49.14 $\pm$ 0.7 & 59.04 $\pm$ 11.37 & 1.0 & 0.5248 & 1.0 & 0.5248 & 4000 & 2099 \\

\midrule

Above Horizon - E1, E2, E3 & 30.58 $\pm$ 3.02 & 50.90 $\pm$ 4.72 & 1.0 & 0.1 & 1.0 & 0.1 & 6000 & 605\\
\cdashmidrule{1-9} 

Below Horizon - E1, E2, E3 & 28.6 $\pm$ 2.73 & 30.46 $\pm$ 3.26 & 1.0 & 0.91 & 1.0 & 0.91  & 6000  & 5466\\
\bottomrule
\end{tabular}
}
\label{tab:isaac_final_v2}
\end{table*}

\begin{table*}
\centering
\caption{Real world benchmarking}
\scalebox{1}{
\begin{tabular}{ccccccccc}
\toprule

& \multicolumn{2}{c}{\textbf{Separation Minima (m) $\uparrow$}} & \multicolumn{2}{c}{\textbf{P(NMAC) $\downarrow$}} & \multicolumn{2}{c}{\textbf{Risk Ratio $\downarrow$}} & \multicolumn{2}{c}{\textbf{Number of violations} $\downarrow$} \\

\cmidrule{2-3} \cmidrule(lr{0.75em}){4-5} \cmidrule(lr{0.75em}){6-7} \cmidrule{8-9}

\textbf{Scenario} & Nominal & \coolname{} & Nominal & \coolname{} & Nominal & \coolname{} & Nominal & \coolname{} \\

\cmidrule{1-1} \cmidrule(lr{0.75em}){2-2} \cmidrule(lr{0.75em}){3-3} \cmidrule(lr{0.75em}){4-4} \cmidrule(lr{0.75em}){5-5} \cmidrule(lr{0.75em}){6-6} \cmidrule(lr{0.75em}){7-7} \cmidrule(lr{0.75em}){8-8} \cmidrule(lr{0.75em}){9-9}

E1 & 41.28 $\pm$ 0.75 & 51.58 $\pm$ 12.76 & 1.0 & 0.33 & 1.0 & 0.5 & 4 & 2 \\

\cdashmidrule{1-9}

E2 & 44.25 $\pm$ 0.64 & 56.22 $\pm$ 2.88 & 1.0 & 0.25 & 1.0 & 0.25 & 4 & 1 \\ 

\cdashmidrule{1-9}

E3 & 47.45 $\pm$ 0.22 & 71.69 $\pm$ 11.65 & 1.0 & 0.5 & 1.0 & 0.25 & 4 & 1 \\

\cdashmidrule{1-9}

E4 & 30.54 $\pm$ 2.07 & 46.05 $\pm$ 8.29 & 1.0 & 0.0 & 1.0 & 0.0 & 2 & 0 \\

\cdashmidrule{1-9}

E5 & 27.39 $\pm$ 5.03 & 38.94 $\pm$ 7.83 & 1.0 & 0.0 & 1.0 & 0.0 & 3 & 0 \\

\midrule

Above Horizon - E1, E2, E3 & 44.43 $\pm$ 3.02 & 59.43 $\pm$ 4.72 & 1.0 & 0.0 & 1.0 & 0.0 & 6  & 0\\
\cdashmidrule{1-9} 

Below Horizon - E1, E2, E3 & 44.59 $\pm$ 2.73 & 60.23 $\pm$ 18.76 & 1.0 & 0.667 & 1.0 & 0.667  & 6  & 4\\
\bottomrule
\end{tabular}
}
\label{tab:real_world_benchmarking}
\end{table*}

\section{\coolname{} Testing}
\label{sec:results}

\subsection{Experiment Design}
We consider a two-agent interaction scenario. The \coolname{}-enabled ego agent is tested against an airborne intruder in various collision geometries. We start by identifying three classes of interaction scenarios for our two-agent setup: Head-on, Overtake, and Lateral. A Head-On scenario has the ego agent and the intruder on a single straight-line converging path moving towards each other. An Overtake scenario has the agents on a single straight-line path with a higher-speed ego agent in the trail of a low-speed intruder. A Lateral scenario has the ego and intruder on perpendicular converging paths.
Additionally, since our visual detection module can be sensitive to the intruder being above or below the horizon, we investigate both possibilities for our agents, leading to 6 total scenarios. Our six configuration setups are described in \cref{fig:collisoions_and_weather}. These experiments are performed in both a high-fidelity digital-twin simulation and real-world settings. Table \ref{tab:exp_setup} shows the various agents, collision geometries, commanded ground speeds, and test locations.

\subsection{\coolname{} Hardware Prototype}

\coolname{} uses custom hardware as shown in \cref{fig:splash}. The hardware has the following major components:
\begin{enumerate}
    \item Cameras: 6 x  Sony IMX264 cameras providing a total
of $220^{\circ}$ FOV horizontally and $48^{\circ}$ vertically.  
\item Compute: NVIDIA AGX Orin Developer Kit
\item Camera Adaptor: Leopard Imaging LI-JXAV-MIPI-ADPT-6CAM-FP
\item State Estimation: 3DM-GQ7 GNS/INSS module with a dual-GPS antenna setup.
\item ADS-B In: uAvionix PING™ MAVLink based ADS-B receiver (not used)
\end{enumerate}
The payload is securely mounted below the ownship on a custom mounting plate. The entire payload weighs approx. 1.6 kgs, including the 3D-printed protective casing. The deep-learning pipeline and the collision avoidance algorithm described in Section \ref{sec:approach} run onboard the payload in real-time. The payload communicates with the ownship autopilot via the MavLink protocol to issue navigation commands in the form of yaw turn rates. We use DroneKit \cite{dronekit_python} to command the ownship autopilot to execute the avoidance maneuvers. The entire system is agnostic to the ownship platform and is designed to be mounted easily on most systems.

\subsection{Reaction Time Profile for Successful Detect \& Avoid}

Based on the empirical performance statistics of our vision inference system (AirTrack \cite{ghosh2023airtrack}), we determine the range profile in which the intruder tracks are reliable and the maximum reaction time available for the \coolname{} system.
Leveraging the empirical numbers from the ASTM F3442/F3442M standards (satisfied by AirTrack using a Bell 407 intruder with a frontal diagonal length of 4.27 m), the minimum number of pixels at which AirTrack can reliably detect the object with a 95\% probability of track comes out to be 14 pixels.
Using the approximate calibrated focal length of our multi-camera payload, the maximum distance $d_{max}$ at which the object can be detected comes out to be $163.2 \times l$, where $l$ is the frontal diagonal length of the intruder.
For our real-world intruder platforms (shown in \cref{fig:splash}), the reliable maximum distance comes out as $d_{max}^{Hexarotor} = 287 m$ and $d_{max}^{VTOL} = 525 m$.
Thus, for real-world Head-on scenarios, the maximum available reaction time for the M600 intruder is $14.35 s$, and for the VTOL intruder, it is $13.12 s$.
This short maximum reaction time signifies the need for accurate, fast, and on-edge DAA, which we find is empirically satisfied by our \coolname{} system as shown in both benchmarking (\cref{tab:isaac_final_v2} and \ref{tab:real_world_benchmarking}) and profiling (\cref{tab:compute_performance}).

\begin{table}[!t]
    \centering
    \caption{ViSafe System Profiling}
    \begin{tabular}{|l|c|}
        \hline
        \vspace{-0.5em} & \vspace{-0.5em} \\
        \textbf{Component} & \textbf{Performance} \\[0.2em]
        \hline
        \vspace{-0.5em} & \vspace{-0.5em} \\
        AirTrack Detection \& Tracking (GPU) & 8 Hz \\
        Multi-View Fusion \& State Estimation (CPU) & 24 Hz \\
        CBF-based Avoidance Control (CPU) & 50 Hz (solving a QP) \\
        Mean CPU Utilization & 55.21~\% \\
        Mean GPU Utilization & 50.02~\% \\
        Peak Memory Consumption & 15.96 GB ~ \\
        \hline
    \end{tabular}
    \label{tab:compute_performance}
\end{table}

\subsection{Simulated Digital-Twin Experiment Setup}

We first perform experiments in a digital twin of the test environment. The digital twin enables low-cost, low-risk testing in a more diverse and controllable setting. 

In order to test the scenario as realistically as possible, we create a digital twin of the real-world testing site. We use the Pegasus Simulator \cite{10556959}, which is a framework built on top of NVIDIA Omniverse and IsaacSim \cite{nvidiaIsaac}. While designing the assets, we use ArcGIS CityEngine \cite{esriProceduralCity}, which takes satellite imagery, terrain maps, and structure information from OpenStreetMap to build a 3D model of the test site, including buildings and roads. We also procedurally added trees and a realistic sky with clouds and weather conditions as depicted in Fig.~\ref{fig:collisoions_and_weather}. The experiments were carried out on a desktop computer with an Nvidia 3080Ti and an Nvidia A6000 running Ubuntu 20.04.

To simulate real-world hardware performance, we transmit the rendered images to the onboard processing units for detection and tracking. This allows us to conduct hardware-in-the-loop (HITL) experiments within the digital-twin framework.
Overall, we simulate 10 weather conditions, including time-of-day and cloud covers, three representative collision configurations, and two altitude setups (Above/Below horizon), leading to a total of 60 distinct scenarios. For the above scenarios, we simulated 200 trajectories per scenario, leading to a total of 12000 trajectories.

\subsection{Real World Experiment Setup}
\label{subsec:real_exp}
We provide details on the real-world experiments to test the \coolname{} system. Real-world flight tests are conducted at two distinct locations. A dedicated testing facility at the Nardo Flight Test Field \cite{he2023foundloc} (Valentin Vassilev Memorial Field), USA \verb|[FAA Identifier: 77PA]|, \verb|[40-35-00.2420N 79-53-59.1890W]| as well as a secondary test-site in Leesburg, VA, USA are utilized for experiments. All tests are performed under the FAA 14 CFR Part 107 guidelines. The Nardo flight test airfield has an unpaved runway with a total length of 1800ft (550m) and a width of 100ft (30.5m). The Leesburg field is an open test range. Overall, we perform two runs for each configuration shown in \cref{tab:exp_setup}. The field experiments use three airborne platforms, as shown in Fig.~\ref{fig:splash}. Their details are as follows:
\begin{enumerate}
    \item Aurelia X6 Standard: A hexa-rotor ownship. 
    \item Hexarotor: A multi-rotor low-speed intruder.
    \item VTOL: A quad-plane vertical takeoff and landing fixed-wing aircraft that acts as our high-speed intruder.
\end{enumerate}

\subsection{Metrics}
We compare our technique with a nominal planner without an avoidance strategy. By default, the nominal plans violated the safe separation clause, and we include it solely as a reference point to quantify ViSafe’s gains via metrics.  Our metrics are inspired by the operational standards proposed for collision avoidance systems by the Radio Technical Commission for Aeronautics (RTCA) \cite{RTCA_DO385A}. These are the key metrics which we use:
\begin{enumerate}
    \item \textit{Probability of Near Mid Air Collision P(NMAC)}:  This metric measures the probability that two agents come within a predefined unsafe distance. Lower values signify better collision avoidance.  
    \item \textit{Separation Minima}: This metric captures the minimum distance between agents during an episode. Higher values indicate greater safety margins. 
    \item \textit{Horizontal Rate of Closure (HROC)}: This metric measures the relative velocity at which two agents approach each other, with negative values indicating impending collisions. Higher values reduce collision risks. 
    \item \textit{Risk Ratio}: This metric measures the ratio of collisions compared to a baseline (nominal planner). Lower values indicates higher safety improvements over the baseline.

\end{enumerate}

\subsection{Results and Discussion}

\begin{figure}
    \includegraphics[width=0.48\textwidth]{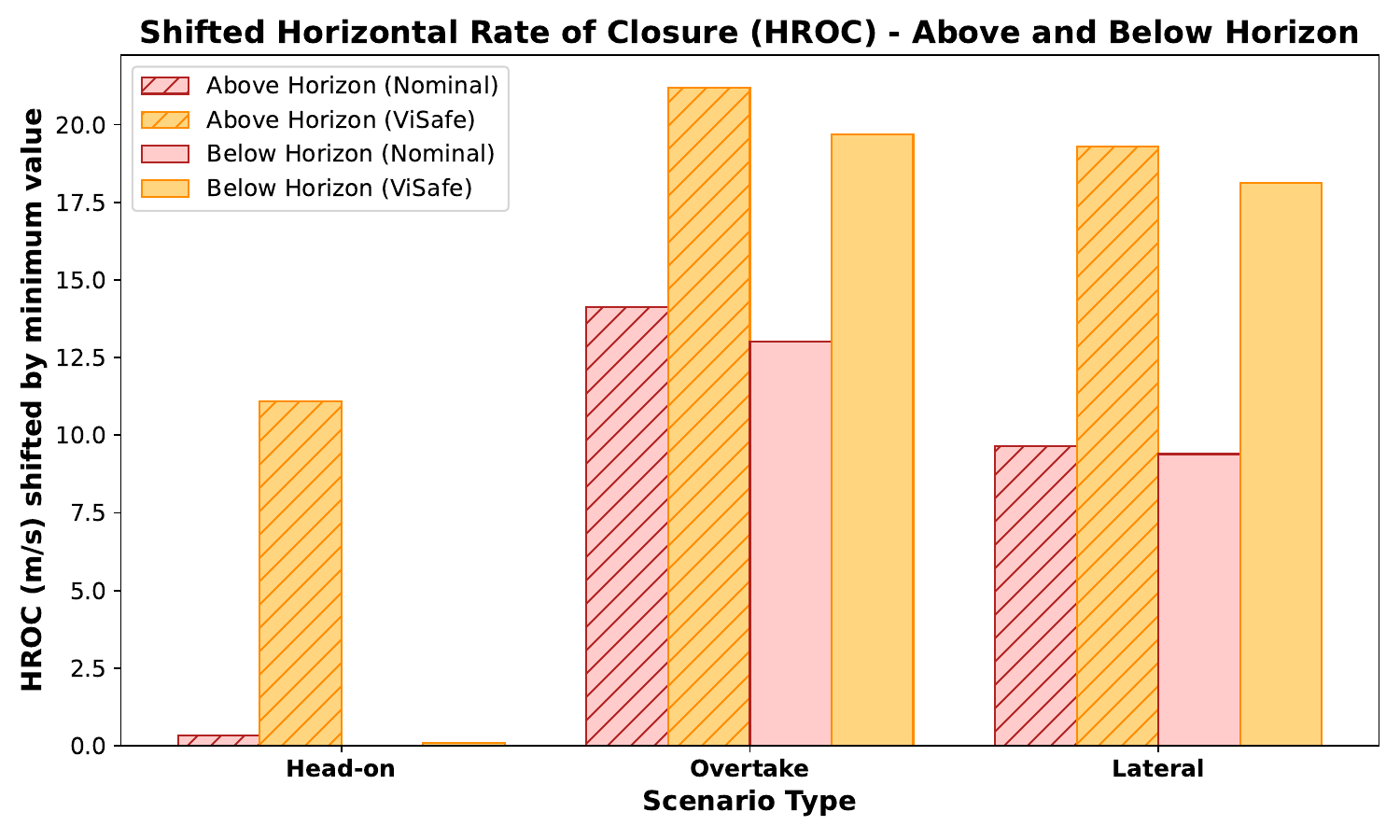}
    \caption{\textbf{Average horizontal rate of closure comparisons across different encounter geometries in real-world testing:} Higher values indicate that agents are moving apart, showcasing diverging \& safe trajectories. Under different testing scenarios, it can be seen that \coolname{} consistently shows a significant boost in average HROC over the nominal plan.}
    \label{fig:hroc}
\end{figure}

\begin{figure*}
    \centering
    \includegraphics[width=0.8\textwidth]{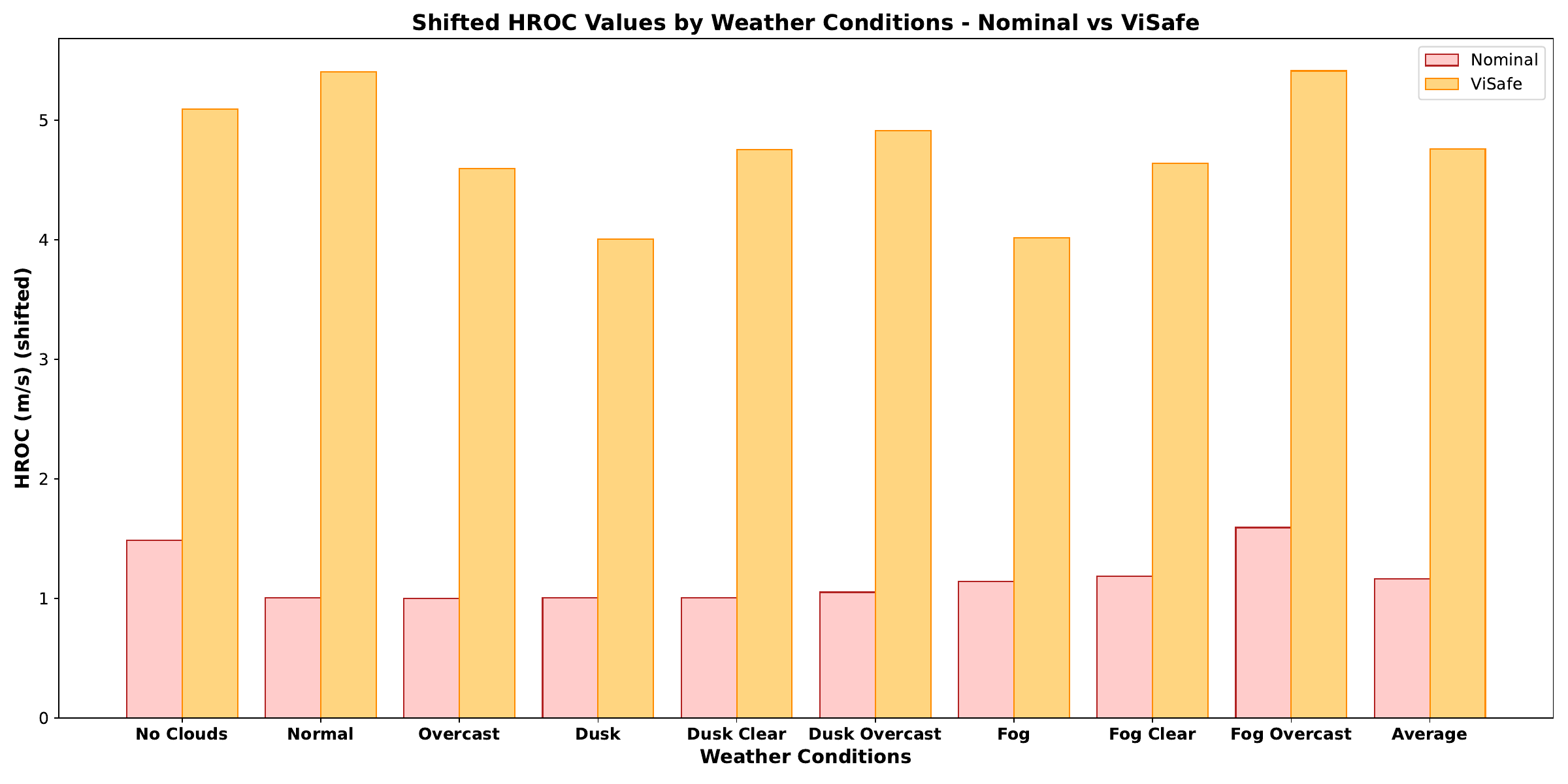}
    \caption{\textbf{Horizontal rate of closure comparisons across different weather conditions in the digital twin:} Higher values indicate that agents are moving apart, showcasing diverging \& safe trajectories. Across the different weather scenarios, \coolname{} showcases consistent boost in HROC over the nominal plan.}
    \label{fig:hroc_sim}
\end{figure*}

In this section, we highlight the key findings of comparing \coolname{} with a nominal planner without an explicit avoidance strategy. The results for simulated digital-twin and real-world experiments are summarized in Tables \ref{tab:isaac_final_v2} and \ref{tab:real_world_benchmarking}.

\subsubsection{\coolname{} enables high-speed collision avoidance \textbf{[H1]}}

Our results demonstrate the advantage of \coolname{} in reducing the collision risk over nominal plan. By design, The nominal planner exhibited high \textit{P(NMAC)} values across Head-On, Overtake, and Lateral configurations for simulation experiments. \coolname{} drastically reduces the \textit{P(NMAC)} for the same configurations. Our method significantly reduces collision risk and improves safety margins, especially in critical configurations such as head-on and overtaking situations. The total number of loss of separations is reduced significantly by \coolname{} for all testing configurations in the real world.

The results in \cref{fig:hroc} indicate that \coolname{} improves the HROC values compared to the nominal planner. In the head-on scenario, HROC improved from \(-19.25 \, \text{m/s}\) (nominal) to \(-13.83 \, \text{m/s}\) (\coolname{}), demonstrating that our method facilitates safer approaches by reducing the rate of closure. Similar trends are observed in lateral and overtaking cases, where \coolname{} maintains a higher and safer HROC, effectively mitigating dangerous approaches.

\subsubsection{\coolname{} is robust to weather and lightning conditions \textbf{[H2]}}
We also test the robustness of \coolname{} to weather and lighting conditions in simulation. The results are highlighted in \ref{tab:isaac_final_v2}. Our risk ratios increase due to detection and tracking failures for some weather conditions. However, our overall risk ratios with respect to all trajectories are still considerably lower than the nominal values across all configurations. \coolname{} also achieves higher HROC values than the nominal ones for all weather conditions, indicating safe \& divergent trajectories as highlighted in \cref{fig:hroc_sim}.

\subsubsection{Digital-twin provides a reliable estimate of real-world performance \textbf{[H3]}} 
Our results indicate that the performance metrics observed in the digital-twin environment closely align with those obtained via real-world testing. As summarized in Tables \ref{tab:isaac_final_v2} and \ref{tab:real_world_benchmarking}, \coolname{} demonstrates higher separation minima and low-risk ratios in all configurations, with slightly higher risk ratios in simulation compared to real-world results.  We hypothesize that this is due to larger simulated data points that capture more edge cases.
In addition, the lower risk ratios observed in real-world tests (indicating higher safety performance) suggest that the digital twin environment provides a conservative estimate of \coolname{}’s performance. These findings underline the importance of the digital twin as a reliable proxy for predeployment evaluation of safety and finding the lower bound. By identifying and remedying edge case behavior in simulation, \coolname{} ensures robust real-world operation.

\section{Learned Challenges and Limitations}

\subsection{Learned Challenges}

There were multiple challenges observed while trying to develop and test \coolname{}. We list a few of them.
\begin{enumerate}
    \item \textbf{Lack of global information availability:} CBF guarantees invariance when information is available throughout the episode. Since we use vision inference to trigger avoidance, the avoidance window is significantly reduced. Theoretical guarantees provided by traditional CBFs can be compromised without making modifications to the design of the CBF itself.
    \item \textbf{Actuation constraints and singularities:} Actuation constraints for the ownship drone proved challenging in the CBF design process. We had to theoretically guarantee safety while ensuring we always had a control output that kept us safe. This was achieved by careful hyperparameter tuning for a specific ownship system. In the future, a more principled approach, such as automatic safety index synthesis \cite{zhao2023safetyindexsynthesissumofsquares}, can be used.
    \item \textbf{Inaccuracies in vision-based inference:} Vision-based state estimation is not perfect; therefore, false positives can often throw the safety module off. We had to use Kalman filtering to ensure minimal false positives were passed, but in the future, we plan to make it explicitly a part of the CBF to ignore out-of-distribution inputs, leveraging Robust CBFs \cite{https://doi.org/10.48550/arxiv.2107.04094}.
    \item \textbf{Safety guarantees:} We acknowledge that the CBF can only provide a formal safety guarantee given always accurate global perception, i.e., no false negatives, which is not the case in the real world.
    However, we believe that ensuring compliance with standards and conducting operability studies supported by empirical statistical analysis (similar to our extensive HITL digital twin testing) can help build valid safety cases for DAA systems.
    \item \textbf{Real world testing constraints:} Testing CBFs on actual drones and collecting data is a time-consuming process, and only a limited number of experiments can be run. Additionally, multiple safety procedures must be put in place before testing, making the process cumbersome. Also, unexpected disturbances like LOS, high wind speeds, and data corruption make testing in the real world challenging. In the future, we plan to use the drone telemetry data we have collected to test our approach and fine-tune it, further exploring ways to improve simulation fidelity.
\end{enumerate}

\subsection{Limitations}

Across our wide array of simulation and real-world tests, we find that our current system struggles when the intruder is below the horizon.
As acknowledged in the benchmarking of our prior work (AirTrack~\cite{ghosh2023airtrack}), the primary reason is that the vision-based inference system fails to establish a consistent track when the background changes fast.
Furthermore, as shown in prior work~\cite{martin2024targeted}, this performance behavior can be attributed to biases or shortcuts learned by the model given the training data distribution.
We aim to leverage optical flow and temporal information in future work to mitigate this issue.
Lastly, for the design of our CBF, we make the following assumptions: the intruder has a constant velocity vector, the action space of the ownship is constrained to a 2D plane parallel to the ground plane, and the intruder doesn't react to the ownship's actions.  Although these are used to make the problem tractable, we aim to investigate interaction-aware avoidance without explicit communication in future
 work using adaptive control barrier functions, improving the versatility of our \coolname{} system.

\section{Conclusion}
\label{sec:conclusion}
In this work, we present a comprehensive solution for high-speed vision-only airborne collision avoidance. 
Our full-stack solution, \coolname{}, deploys perceptual input-focused control barrier functions (CBF) for safe self-separation in dynamic environments. 
Through simulated hardware-in-the-loop digital twin testing and real-world flight tests, we exemplify \coolname{}'s reliability in the face of heterogeneous agents, varying encounter geometries, and other environmental factors. 
\coolname{}’s successful deployment in real-world high-speed collision avoidance tests (at closure rates of up to 144 km/h) establishes its effectiveness as a cornerstone for autonomous aerial navigation systems.

\section*{Acknowledgments}

Approved for public release; distribution is unlimited.
This research was sponsored by DARPA (W911NF-18-2-0218).
The views, opinions, and/or findings expressed are those of the author(s) and should not be interpreted as representing the official views or policies of the Department of Defense or the U.S. Government.
The authors thank Rebecca Martin, Helen Wang, Nayana Suvarna, and folks at Near Earth Autonomy for their support with the field testing of ViSafe. 
We also thank John Keller for helping out with the payload.


\begin{thebibliography}{57}
\providecommand{\natexlab}[1]{#1}
\providecommand{\url}[1]{\texttt{#1}}
\expandafter\ifx\csname urlstyle\endcsname\relax
  \providecommand{\doi}[1]{doi: #1}\else
  \providecommand{\doi}{doi: \begingroup \urlstyle{rm}\Url}\fi

\bibitem[air()]{aircrowd}
Airborne object tracking dataset.
\newblock URL \url{https://registry.opendata.aws/airborne-object-tracking}.

\bibitem[dro()]{dronekit_python}
Dronekit-python: Developer sdk for python to control drones using the mavlink protocol.
\newblock URL \url{https://github.com/dronekit/dronekit-python}.

\bibitem[esr()]{esriProceduralCity}
{P}rocedural {C}ity {G}enerator | 3{D} {C}ity {M}aker | {A}rc{G}{I}{S} {C}ity{E}ngine --- esri.com.
\newblock \url{https://www.esri.com/en-us/arcgis/products/arcgis-cityengine/overview}.
\newblock [Accessed 11-12-2024].

\bibitem[nvi()]{nvidiaIsaac}
{I}saac {S}im --- developer.nvidia.com.
\newblock \url{https://developer.nvidia.com/isaac/sim}.
\newblock [Accessed 11-12-2024].

\bibitem[isa(2022)]{isaac_sim_ref}
Nvidia isaac sim, 2022.
\newblock URL \url{https://developer.nvidia.com/isaac-sim}.

\bibitem[Agrawal and Sreenath(2017)]{Agrawal2017DiscreteCB}
Ayush Agrawal and Koushil Sreenath.
\newblock Discrete control barrier functions for safety-critical control of discrete systems with application to bipedal robot navigation.
\newblock In \emph{Robotics: Science and Systems}, 2017.
\newblock URL \url{https://api.semanticscholar.org/CorpusID:1780280}.

\bibitem[Ames et~al.(2019)Ames, Coogan, Egerstedt, Notomista, Sreenath, and Tabuada]{8796030}
Aaron~D. Ames, Samuel Coogan, Magnus Egerstedt, Gennaro Notomista, Koushil Sreenath, and Paulo Tabuada.
\newblock Control barrier functions: Theory and applications.
\newblock In \emph{2019 18th European Control Conference (ECC)}, pages 3420--3431, 2019.
\newblock \doi{10.23919/ECC.2019.8796030}.

\bibitem[Asmar et~al.(2013)Asmar, Kochenderfer, Asmar, and Kochenderfer]{asmar2013optimized}
DM~Asmar, MJ~Kochenderfer, DM~Asmar, and MJ~Kochenderfer.
\newblock Optimized airborne collision avoidance in mixed equipage environments.
\newblock \emph{Lincoln Laboratory, MIT Report}, 2013.

\bibitem[Bewley et~al.(2016)Bewley, Ge, Ott, Ramos, and Upcroft]{bewley2016simple}
Alex Bewley, Zongyuan Ge, Lionel Ott, Fabio Ramos, and Ben Upcroft.
\newblock Simple online and realtime tracking.
\newblock In \emph{2016 IEEE international conference on image processing (ICIP)}, pages 3464--3468. IEEE, 2016.

\bibitem[Breeden and Panagou(2021)]{https://doi.org/10.48550/arxiv.2107.04094}
Joseph Breeden and Dimitra Panagou.
\newblock Robust control barrier functions under high relative degree and input constraints for satellite trajectories, 2021.
\newblock URL \url{https://arxiv.org/abs/2107.04094}.

\bibitem[Carnie et~al.(2006)Carnie, Walker, and Corke]{carnie2006image}
Ryan Carnie, Rodney Walker, and Peter Corke.
\newblock Image processing algorithms for uav" sense and avoid".
\newblock In \emph{Proceedings 2006 IEEE International Conference on Robotics and Automation, 2006. ICRA 2006.}, pages 2848--2853. IEEE, 2006.

\bibitem[Chen et~al.(2014)Chen, Xiang, Liu, and Pan]{chen2014aircraft}
Xueyun Chen, Shiming Xiang, Cheng-Lin Liu, and Chun-Hong Pan.
\newblock Aircraft detection by deep convolutional neural networks.
\newblock \emph{IPSJ Transactions on Computer Vision and Applications}, 7:\penalty0 10--17, 2014.

\bibitem[Dey et~al.(2010)Dey, Geyer, Singh, and Digioia]{dey2010passive}
Debadeepta Dey, Christopher Geyer, Sanjiv Singh, and Matt Digioia.
\newblock Passive, long-range detection of aircraft: towards a field deployable sense and avoid system.
\newblock In \emph{Field and Service Robotics}, pages 113--123. Springer, 2010.

\bibitem[Duan et~al.(2019)Duan, Bai, Xie, Qi, Huang, and Tian]{duan2019centernet}
Kaiwen Duan, Song Bai, Lingxi Xie, Honggang Qi, Qingming Huang, and Qi~Tian.
\newblock Centernet: Keypoint triplets for object detection.
\newblock In \emph{Proceedings of the IEEE/CVF international conference on computer vision}, pages 6569--6578, 2019.

\bibitem[Fasano et~al.(2014)Fasano, Accardo, Tirri, Moccia, and De~Lellis]{fasano2014morphological}
Giancarmine Fasano, Domenico Accardo, Anna~Elena Tirri, Antonio Moccia, and Ettore De~Lellis.
\newblock Morphological filtering and target tracking for vision-based uas sense and avoid.
\newblock In \emph{2014 International Conference on Unmanned Aircraft Systems (ICUAS)}, pages 430--440. IEEE, 2014.

\bibitem[Fu et~al.(2023)Fu, Wen, and Cao]{10327509}
Junjie Fu, Guanghui Wen, and Jinde Cao.
\newblock High-order control barrier function based collision avoidance formation tracking of constrained fixed-wing aircraft.
\newblock In \emph{2023 5th International Conference on Industrial Artificial Intelligence (IAI)}, pages 1--6, 2023.
\newblock \doi{10.1109/IAI59504.2023.10327509}.

\bibitem[Ghosh et~al.(2023)Ghosh, Patrikar, Moon, Hamidi, and Scherer]{ghosh2023airtrack}
Sourish Ghosh, Jay Patrikar, Brady Moon, Milad~Moghassem Hamidi, and Sebastian Scherer.
\newblock Airtrack: Onboard deep learning framework for long-range aircraft detection and tracking.
\newblock In \emph{2023 IEEE International Conference on Robotics and Automation (ICRA)}, pages 1277--1283, 2023.
\newblock \doi{10.1109/ICRA48891.2023.10160627}.

\bibitem[Hamissi et~al.(2024)Hamissi, Dhraief, and Sliman]{hamissi2024comprehensive}
Asma Hamissi, Amine Dhraief, and Layth Sliman.
\newblock A comprehensive survey on conflict detection and resolution in unmanned aircraft system traffic management.
\newblock \emph{IEEE Transactions on Intelligent Transportation Systems}, 2024.

\bibitem[Harms et~al.(2024)Harms, Kulkarni, Khedekar, Jacquet, and Alexis]{harms2024neuralcontrolbarrierfunctions}
Marvin Harms, Mihir Kulkarni, Nikhil Khedekar, Martin Jacquet, and Kostas Alexis.
\newblock Neural control barrier functions for safe navigation, 2024.
\newblock URL \url{https://arxiv.org/abs/2407.19907}.

\bibitem[He et~al.(2023)He, Cisneros, Keetha, Patrikar, Ye, Higgins, Hu, Kapoor, and Scherer]{he2023foundloc}
Yao He, Ivan Cisneros, Nikhil Keetha, Jay Patrikar, Zelin Ye, Ian Higgins, Yaoyu Hu, Parv Kapoor, and Sebastian Scherer.
\newblock Foundloc: Vision-based onboard aerial localization in the wild.
\newblock \emph{arXiv preprint arXiv:2310.16299}, 2023.

\bibitem[Hu and Ramanan(2017)]{hu2017finding}
Peiyun Hu and Deva Ramanan.
\newblock Finding tiny faces.
\newblock In \emph{Proceedings of the IEEE conference on computer vision and pattern recognition}, pages 951--959, 2017.

\bibitem[Hwang et~al.(2018)Hwang, Lee, Shin, Cho, and Shim]{hwang2018aircraft}
Sunyou Hwang, Jaehyun Lee, Heemin Shin, Sungwook Cho, and David~Hyunchul Shim.
\newblock Aircraft detection using deep convolutional neural network in small unmanned aircraft systems.
\newblock In \emph{2018 AIAA Information Systems-AIAA Infotech@ Aerospace}, page 2137, 2018.

\bibitem[Jacinto et~al.(2024)Jacinto, Pinto, Patrikar, Keller, Cunha, Scherer, and Pascoal]{10556959}
Marcelo Jacinto, João Pinto, Jay Patrikar, John Keller, Rita Cunha, Sebastian Scherer, and António Pascoal.
\newblock Pegasus simulator: An isaac sim framework for multiple aerial vehicles simulation.
\newblock In \emph{2024 International Conference on Unmanned Aircraft Systems (ICUAS)}, pages 917--922, 2024.
\newblock \doi{10.1109/ICUAS60882.2024.10556959}.

\bibitem[James et~al.(2018)James, Ford, and Molloy]{james2018learning}
Jasmin James, Jason~J Ford, and Timothy~L Molloy.
\newblock Learning to detect aircraft for long-range vision-based sense-and-avoid systems.
\newblock \emph{IEEE Robotics and Automation Letters}, 3\penalty0 (4):\penalty0 4383--4390, 2018.

\bibitem[James et~al.(2019)James, Ford, and Molloy]{james2019below}
Jasmin James, Jason~J Ford, and Timothy~L Molloy.
\newblock Below horizon aircraft detection using deep learning for vision-based sense and avoid.
\newblock In \emph{2019 International Conference on Unmanned Aircraft Systems (ICUAS)}, pages 965--970. IEEE, 2019.

\bibitem[Keetha et~al.(2022)Keetha, Wang, Qiu, Xu, and Scherer]{keetha2022airobject}
Nikhil~Varma Keetha, Chen Wang, Yuheng Qiu, Kuan Xu, and Sebastian Scherer.
\newblock Airobject: A temporally evolving graph embedding for object identification.
\newblock In \emph{Proceedings of the IEEE/CVF Conference on Computer Vision and Pattern Recognition}, pages 8407--8416, 2022.

\bibitem[Kuchar and Drumm(2007)]{kuchar2007traffic}
JE~Kuchar and Ann~C Drumm.
\newblock The traffic alert and collision avoidance system.
\newblock \emph{Lincoln laboratory journal}, 16\penalty0 (2):\penalty0 277, 2007.

\bibitem[Lai et~al.(2008)Lai, Ford, O'Shea, and Walker]{lai2008hidden}
John Lai, Jason~J Ford, Peter O'Shea, and Rodney Walker.
\newblock Hidden markov model filter banks for dim target detection from image sequences.
\newblock In \emph{2008 Digital Image Computing: Techniques and Applications}, pages 312--319. IEEE, 2008.

\bibitem[Lai et~al.(2011)Lai, Mejias, and Ford]{lai2011airborne}
John Lai, Luis Mejias, and Jason~J Ford.
\newblock Airborne vision-based collision-detection system.
\newblock \emph{Journal of Field Robotics}, 28\penalty0 (2):\penalty0 137--157, 2011.

\bibitem[Lai et~al.(2013)Lai, Ford, Mejias, and O'Shea]{lai2013characterization}
John Lai, Jason~J Ford, Luis Mejias, and Peter O'Shea.
\newblock Characterization of sky-region morphological-temporal airborne collision detection.
\newblock \emph{Journal of Field Robotics}, 30\penalty0 (2):\penalty0 171--193, 2013.

\bibitem[Liang et~al.(2021)Liang, Wei, Zhang, Geng, Zhang, Sun, Zhou, Wei, and Gao]{liang2021learning}
Dong Liang, Zongqi Wei, Dong Zhang, Qixiang Geng, Liyan Zhang, Han Sun, Huiyu Zhou, Mingqiang Wei, and Pan Gao.
\newblock Learning calibrated-guidance for object detection in aerial images.
\newblock \emph{arXiv preprint arXiv:2103.11399}, 2021.

\bibitem[Liu and Tomizuka(2014)]{Liu2014CONTROLIA}
Changliu Liu and Masayoshi Tomizuka.
\newblock Control in a safe set: Addressing safety in human-robot interactions.
\newblock In \emph{IEEE/ACM International Conference on Human-Robot Interaction}, 2014.

\bibitem[Manfredi and Jestin(2016)]{7778055}
Guido Manfredi and Yannick Jestin.
\newblock An introduction to acas xu and the challenges ahead.
\newblock In \emph{2016 IEEE/AIAA 35th Digital Avionics Systems Conference (DASC)}, pages 1--9, 2016.
\newblock \doi{10.1109/DASC.2016.7778055}.

\bibitem[Martin et~al.(2024)Martin, Fung, Keetha, Bauer, and Scherer]{martin2024targeted}
Rebecca Martin, Clement Fung, Nikhil Keetha, Lujo Bauer, and Sebastian Scherer.
\newblock Targeted image transformation for improving robustness in long range aircraft detection.
\newblock In \emph{2024 IEEE/RSJ International Conference on Intelligent Robots and Systems (IROS)}, pages 10431--10438. IEEE, 2024.

\bibitem[McCandless(1999)]{mccandless1999detection}
Jeffrey~W McCandless.
\newblock Detection of aircraft in video sequences using a predictive optical flow algorithm.
\newblock \emph{Optical Engineering}, 38\penalty0 (3):\penalty0 523--530, 1999.

\bibitem[Mejias et~al.(2010)Mejias, McNamara, Lai, and Ford]{mejias2010vision}
Luis Mejias, Scott McNamara, John Lai, and Jason Ford.
\newblock Vision-based detection and tracking of aerial targets for uav collision avoidance.
\newblock In \emph{2010 IEEE/RSJ International Conference on Intelligent Robots and Systems}, pages 87--92. IEEE, 2010.

\bibitem[Molloy et~al.(2017)Molloy, Ford, and Mejias]{molloy2017detection}
Timothy~L Molloy, Jason~J Ford, and Luis Mejias.
\newblock Detection of aircraft below the horizon for vision-based detect and avoid in unmanned aircraft systems.
\newblock \emph{Journal of Field Robotics}, 34\penalty0 (7):\penalty0 1378--1391, 2017.

\bibitem[Molnar et~al.(2024)Molnar, Kannan, Cunningham, Dunlap, Hobbs, and Ames]{molnar2024collisionavoidancegeofencingfixedwing}
Tamas~G. Molnar, Suresh~K. Kannan, James Cunningham, Kyle Dunlap, Kerianne~L. Hobbs, and Aaron~D. Ames.
\newblock Collision avoidance and geofencing for fixed-wing aircraft with control barrier functions, 2024.
\newblock URL \url{https://arxiv.org/abs/2403.02508}.

\bibitem[Murtaza et~al.(2021)Murtaza, Aguilera, Azimi, and Hutchinson]{9636794}
Muhammad~Ali Murtaza, Sergio Aguilera, Vahid Azimi, and Seth Hutchinson.
\newblock Real-time safety and control of robotic manipulators with torque saturation in operational space.
\newblock In \emph{2021 IEEE/RSJ International Conference on Intelligent Robots and Systems (IROS)}, pages 702--708, 2021.
\newblock \doi{10.1109/IROS51168.2021.9636794}.

\bibitem[Nussberger et~al.(2014)Nussberger, Grabner, and Van~Gool]{nussberger2014aerial}
Andreas Nussberger, Helmut Grabner, and Luc Van~Gool.
\newblock Aerial object tracking from an airborne platform.
\newblock In \emph{2014 international conference on unmanned aircraft systems (ICUAS)}, pages 1284--1293. IEEE, 2014.

\bibitem[Owen et~al.(2014)Owen, Duffy, and Edwards]{6875652}
Michael~P. Owen, Sean~M. Duffy, and Matthew W.~M. Edwards.
\newblock Unmanned aircraft sense and avoid radar: Surrogate flight testing performance evaluation.
\newblock In \emph{2014 IEEE Radar Conference}, pages 0548--0551, 2014.
\newblock \doi{10.1109/RADAR.2014.6875652}.

\bibitem[Patrikar et~al.(2022)Patrikar, Dantas, Ghosh, Kapoor, Higgins, Aloor, Navarro, Sun, Stoler, Hamidi, et~al.]{patrikar2022challenges}
Jay Patrikar, Joao Dantas, Sourish Ghosh, Parv Kapoor, Ian Higgins, Jasmine~J Aloor, Ingrid Navarro, Jimin Sun, Ben Stoler, Milad Hamidi, et~al.
\newblock Challenges in close-proximity safe and seamless operation of manned and unmanned aircraft in shared airspace.
\newblock \emph{arXiv preprint arXiv:2211.06932}, 2022.

\bibitem[Patrikar et~al.(2025)Patrikar, Dantas, Moon, Hamidi, Ghosh, Keetha, Higgins, Chandak, Yoneyama, and Scherer]{patrikar2025image}
Jay Patrikar, Joao Dantas, Brady Moon, Milad Hamidi, Sourish Ghosh, Nikhil Keetha, Ian Higgins, Atharva Chandak, Takashi Yoneyama, and Sebastian Scherer.
\newblock Image, speech, and ads-b trajectory datasets for terminal airspace operations.
\newblock \emph{Scientific Data}, 12\penalty0 (1):\penalty0 468, 2025.

\bibitem[Petridis et~al.(2008)Petridis, Geyer, and Singh]{petridis2008learning}
Stavros Petridis, Christopher Geyer, and Sanjiv Singh.
\newblock Learning to detect aircraft at low resolutions.
\newblock In \emph{International Conference on Computer Vision Systems}, pages 474--483. Springer, 2008.

\bibitem[Reilly et~al.(2010)Reilly, Idrees, and Shah]{reilly2010detection}
Vladimir Reilly, Haroon Idrees, and Mubarak Shah.
\newblock Detection and tracking of large number of targets in wide area surveillance.
\newblock In \emph{European conference on computer vision}, pages 186--199. Springer, 2010.

\bibitem[Rozantsev et~al.(2015)Rozantsev, Lepetit, and Fua]{rozantsev2015flying}
Artem Rozantsev, Vincent Lepetit, and Pascal Fua.
\newblock Flying objects detection from a single moving camera.
\newblock In \emph{Proceedings of the IEEE Conference on Computer Vision and Pattern Recognition}, pages 4128--4136, 2015.

\bibitem[{RTCA}(2023)]{RTCA_DO385A}
{RTCA}.
\newblock \emph{{Minimum Operational Performance Standards for Airborne Collision Avoidance System X (ACAS X) (ACAS Xa and ACAS Xo), Volume I and Volume II}}.
\newblock DO 385A. RTCA, 2023.
\newblock Approved by the RTCA Program Management Committee (PMC) as documented in June 2023.

\bibitem[Schubert and Mikolajczyk(2014)]{schubert2014robust}
Falk Schubert and Krystian Mikolajczyk.
\newblock Robust registration and filtering for moving object detection in aerial videos.
\newblock In \emph{2014 22nd International Conference on Pattern Recognition}, pages 2808--2813. IEEE, 2014.

\bibitem[Shi et~al.(2023)Shi, Gong, Jiang, Zhi, Bao, Sun, and Zhang]{shi2023complex}
Tianjun Shi, Jinnan Gong, Shikai Jiang, Xiyang Zhi, Guangzhen Bao, Yu~Sun, and Wei Zhang.
\newblock Complex optical remote-sensing aircraft detection dataset and benchmark.
\newblock \emph{IEEE Transactions on Geoscience and Remote Sensing}, 61:\penalty0 1--9, 2023.

\bibitem[Smyers et~al.(2023)Smyers, Katz, Corso, and Kochenderfer]{smyers2023avoidds}
Elysia Smyers, Sydney Katz, Anthony Corso, and Mykel~J Kochenderfer.
\newblock Avoidds: Aircraft vision-based intruder detection dataset and simulator.
\newblock \emph{Advances in Neural Information Processing Systems}, 36:\penalty0 8076--8091, 2023.

\bibitem[Squires et~al.(2018)Squires, Pierpaoli, and Egerstedt]{8511342}
Eric Squires, Pietro Pierpaoli, and Magnus Egerstedt.
\newblock Constructive barrier certificates with applications to fixed-wing aircraft collision avoidance.
\newblock In \emph{2018 IEEE Conference on Control Technology and Applications (CCTA)}, pages 1656--1661, 2018.
\newblock \doi{10.1109/CCTA.2018.8511342}.

\bibitem[Squires et~al.(2019)Squires, Pierpaoli, Konda, Coogan, and Egerstedt]{DBLP:journals/corr/abs-1906-03771}
Eric Squires, Pietro Pierpaoli, Rohit Konda, Samuel Coogan, and Magnus Egerstedt.
\newblock Composition of safety constraints with applications to decentralized fixed-wing collision avoidance.
\newblock \emph{CoRR}, abs/1906.03771, 2019.
\newblock URL \url{http://arxiv.org/abs/1906.03771}.

\bibitem[Stojni{\'c} et~al.(2021)Stojni{\'c}, Risojevi{\'c}, Mu{\v{s}}tra, Jovanovi{\'c}, Filipi, Kezi{\'c}, and Babi{\'c}]{stojnic2021method}
Vladan Stojni{\'c}, Vladimir Risojevi{\'c}, Mario Mu{\v{s}}tra, Vedran Jovanovi{\'c}, Janja Filipi, Nikola Kezi{\'c}, and Zdenka Babi{\'c}.
\newblock A method for detection of small moving objects in uav videos.
\newblock \emph{Remote Sensing}, 13\penalty0 (4):\penalty0 653, 2021.

\bibitem[Yang et~al.(2024)Yang, Tian, Zhao, Wang, Luo, Pu, Zhou, and Pi]{yang2024robust}
Bo~Yang, Dongjian Tian, Songliang Zhao, Wei Wang, Jun Luo, Huayan Pu, Mingliang Zhou, and Yangjun Pi.
\newblock Robust aircraft detection in imbalanced and similar classes with a multi-perspectives aircraft dataset.
\newblock \emph{IEEE Transactions on Intelligent Transportation Systems}, 2024.

\bibitem[Yu et~al.(2024)Yu, Yu, Naddaf-Sh, Upadhyay, Gao, and Fan]{yu2024efficientmotionplanningmanipulators}
Mingxin Yu, Chenning Yu, M-Mahdi Naddaf-Sh, Devesh Upadhyay, Sicun Gao, and Chuchu Fan.
\newblock Efficient motion planning for manipulators with control barrier function-induced neural controller, 2024.
\newblock URL \url{https://arxiv.org/abs/2404.01184}.

\bibitem[Zhao et~al.(2023)Zhao, He, Wei, Liu, and Liu]{zhao2023safetyindexsynthesissumofsquares}
Weiye Zhao, Tairan He, Tianhao Wei, Simin Liu, and Changliu Liu.
\newblock Safety index synthesis via sum-of-squares programming, 2023.
\newblock URL \url{https://arxiv.org/abs/2209.09134}.

\bibitem[Zhou et~al.(2020)Zhou, Koltun, and Kr{\"a}henb{\"u}hl]{zhou2020tracking}
Xingyi Zhou, Vladlen Koltun, and Philipp Kr{\"a}henb{\"u}hl.
\newblock Tracking objects as points.
\newblock In \emph{European Conference on Computer Vision}, pages 474--490. Springer, 2020.

\end{thebibliography}

\end{document}